# 2-speed network ensemble for efficient classification of incremental land-use/land-cover satellite image chips

**Michael James Horry**[a,b], **Subrata Chakraborty**[c,a*], **Biswajeet Pradhan**[a,d,e], **Nagesh Shukla**[f] and **Sanjoy Paul**[g]

[a] Centre for Advanced Modelling and Geospatial Information Systems (CAMGIS), Faculty of Engineering and Information Technology, University of Technology Sydney, Ultimo, NSW 2007, Australia

[b] IBM Australia Limited, Sydney, NSW 2000, Australia

[c] School of Science and Technology, Faculty of Science, Agriculture, Business and Law, University of New England, Armidale, NSW 2351, Australia

[d] Center of Excellence for Climate Change Research, King Abdulaziz University, Jeddah 21589, Saudi Arabia

[e] Earth Observation Center, Institute of Climate Change, Universiti Kebangsaan Malaysia, Selangor 43600, Bangi, Malaysia

[f] School of Professional Practice and Leadership, University of Technology Sydney, Ultimo, NSW 2007, Australia

[g] UTS Business School, University of Technology Sydney, Ultimo, NSW 2007, Australia

* Correspondence: Subrata.Chakraborty@une.edu.au

**Abstract:** The ever-growing volume of satellite imagery data presents a challenge for industry and governments making data-driven decisions based on the timely analysis of very large data sets. Commonly used deep learning algorithms for automatic classification of satellite images are time and resource-intensive to train. The cost of retraining in the context of "Big Data" presents a practical challenge when new image data and/or classes are added to a training corpus. Recognizing the need for an adaptable, accurate, and scalable satellite image chip classification scheme, in this research we present an ensemble of: i) a slow to train but high accuracy vision transformer; and ii) a fast to train, low-parameter convolutional neural network. The vision transformer model provides a scalable and accurate foundation model. The high-speed CNN provides an efficient means of incorporating newly labelled data into analysis, at the expense of lower accuracy. To simulate incremental data, the very large (~400,000 images) So2Sat LCZ42 satellite image chip dataset is divided into four intervals, with the high-speed CNN retrained every interval and the vision transformer trained every half interval. This experimental setup mimics an increase in data volume and diversity over time. For the task of automated land-cover/land-use classification, the ensemble models for each data increment outperform each of the component models, with best accuracy of 65% against a holdout test partition of the So2Sat dataset. The proposed ensemble and staggered training schedule provide a scalable and cost-effective satellite image classification scheme that is optimized to process very large volumes of satellite data.

## 1. Introduction

From the first Landsat mission launched on 23 July 1972, industry, intelligence, and policy-making bodies have used satellite imagery data as a primary source of information relating to land-use and land-cover change [1]. Initially, automated methods for the analysis of this data were based on pixel analysis since the coarse-grained image pixels contained features of interest within the pixel boundary [2], and comparative changes of discrete pixel values within the same image chip could be used to indicate land-use/land cover changes [3]. Modern satellite images comprise pixels much finer than typical objects of interest. Concurrently, computational methods for the analysis of satellite images have evolved from hand-crafted feature extraction techniques such as histogram analysis, Gist [4], scale-invariant feature transform (SIFT) [5], and histogram of oriented gradients (HOG) [6], through unsupervised machine learning methods such as principal component analysis (PCA) [7], Random Forest [8], Support Vector Machines (SVM) [9], and K-Means clustering [10], to supervised deep learning systems. Deep learning systems combine automatic feature extraction and classification using multi-layer neural networks, typically variants of the convolutional neural network (CNN) [11].

An extensive study into the comparative performance of deep learning algorithms against hand-crafted feature extraction in the context of a large and diverse satellite image dataset was performed by [12]. This study noted that handcrafted methods were typically evaluated against small datasets, resulting in an unknown performance at scale. They evaluated hand-crafted, unsupervised learning, and deep learning algorithms against a large dataset (NWPU-RESISC45) consisting of 31,500 high-resolution image samples with an even distribution over 45 scene classes [13]. It was shown that deep learning CNN models outperformed (on accuracy metrics) the tested handcrafted and unsupervised learning algorithms by a margin of at least 30%. A further performance boost of over 6% was achieved by fine-tuning off the shelf CNN models with VGG-16 [14] achieving the highest accuracy for this task of over 90%. For comparison, none of the tested hand-crafted or unsupervised learning algorithms in this study achieved accuracy greater than 45%.

Training very deep neural networks such as CNNs is time-consuming and resource-intensive since multiple passes through large volumes of training data are needed to establish the optimum parameter values for millions of neurons in the network [15]. This has resulted in several researchers calling attention to the energy consumed in training and retraining these models [16, 17]. Incorporating new data and/or class labels into a deep learning model requires the model to be either fully retrained on a revised release of the entire training data corpus, which is time-consuming and resource-intensive, or fine-tuned with new training data as new samples become available. Unfortunately, fine-tuning using a limited set of new samples may lead to overfitting the model to new sample data unless great care is taken to appropriately weight these samples during the training process. For a model initially trained on a sufficiently large data corpus, a small fine-tuning corpus may be weighted so low as to be inconsequential, frustrating the fine-tuning effort and rendering the retraining exercise futile.

A second complication encountered in fine-tuning pre-trained models is that new data may also bring new labels requiring a revised deep learning network architecture with the number of output neurons matching the revised number of class labels. In this case, a transfer learning approach [18] is not feasible, since this requires a match between the neural network architecture of the source and target models. Knowledge distillation [19] techniques using the teacher/student paradigm provide a means of incorporating limited new data into "student" models [20], but these methods are in their infancy and not proven at scale [21, 22]. Although adaptable to new data, one practical hindrance to adopting teacher/student models is the management complexities of large numbers of resultant specialized student models and the question of how student model label scores are best combined into a domain level prediction.

Addressing these limitations of existing approaches to incorporating new land-use/landcover image data and related class labels in deep learning models, this study introduces a two-speed ensemble model consisting of a low parameter, high-speed convolutional neural network (HS-CNN) combined with a highly scalable and accurate but slower to train Vision Transformer (ViT) [23]. We show that this ensemble can be used to produce a scalable, accurate, and adaptable computer vision model for satellite image chip classification of land-use/land-cover as represented by standard climate zone labels. Although the presented technique can be applied to any "big data" computer vision task, satellite image data was selected as the target domain of this study due to the need to solve the real-world problem of efficiently processing the ever-increasing volume and velocity of remote sensing and geospatial data resulting from the continuous growth of open data access missions such as Landsat, Sentinel-1, Sentinel-2, Sentinel-3, MODIS, SRTM and ASTER [24].

## 2. Materials and Methods

The claimed superior performance of the ViT architecture over the CNN architecture is emergent only for very large datasets [23, 25]. Therefore, the proposed two-speed ensemble is trained against the very large So2Sat LCZ42 (So2Sat) dataset consisting of approximately 400,000 image chips sourced from the Sentinel-1 and Sentinel-2 missions [26]. Compared to other frequently cited sources of satellite data So2Sat is an order of magnitude larger as shown in Table 1.

**Table 1.** Summary of publicly available satellite and aerial image dataset considered for use in this study

| Dataset | Source | Labels | Number of Images | Dimensions |
|---|---|---|---|---|
| So2Sat [26] | Satellite | 17 | 400,673 | 32 x 32 |
| UC-Merced [27] | Aerial | 21 | 2,100 | 256 x 256 |
| AID [28] | Aerial | 31 | 10,000 | 600 x 600 |
| Optimal31 [29] | Aerial | 31 | 1,860 | 256 x 256 |
| NWPU-45 [12] | Aerial | 45 | 31,500 | 256 x 256 |
| WHU-RS19 [30] | Satellite | 19 | 1,005 | 600 x 600 |
| RSSCN7 [31] | Satellite | 7 | 2,800 | 400 x 400 |
| SIRI-WHU [32] | Satellite | 12 | 2,400 | 200 x 200 |

This study is concerned with incorporating incremental data into deep learning models. New data acquisition is simulated by splitting the static So2Sat dataset into 25% increments representing four points in time. The 25% data split represents a point where only the HS-CNN is fully trained. Therefore, the classification model for this smallest data segment is the HS-CNN alone. At the 50% data split the ViT and the HS-CNN are both trained with 50% of the total data. At the 75% split the ViT is still trained with only 50% of the full data and the HS-CNN is trained with 75% of the data, representing the real-world experience of new data of 25% which can be rapidly included in the HS-CNN, but not the slower to train ViT. At the 100% data split, both the ViT and the HS-CNN are trained on all available data. This experiment flow is depicted in Figure 1.

It is envisaged that in a real-world implementation, this process of staggered training would continue indefinitely, using retraining of the HS-CNN to rapidly incorporate new labelled data into the ensemble model whilst the ViT "catches-up" at a slower speed. For example, the HS-CNN could be trained hourly with the ViT trained on a daily or weekly basis. The most effective schedule would be determined empirically in the context of empirical factors such as the rate of new data acquisition and compute resource availability and cost.

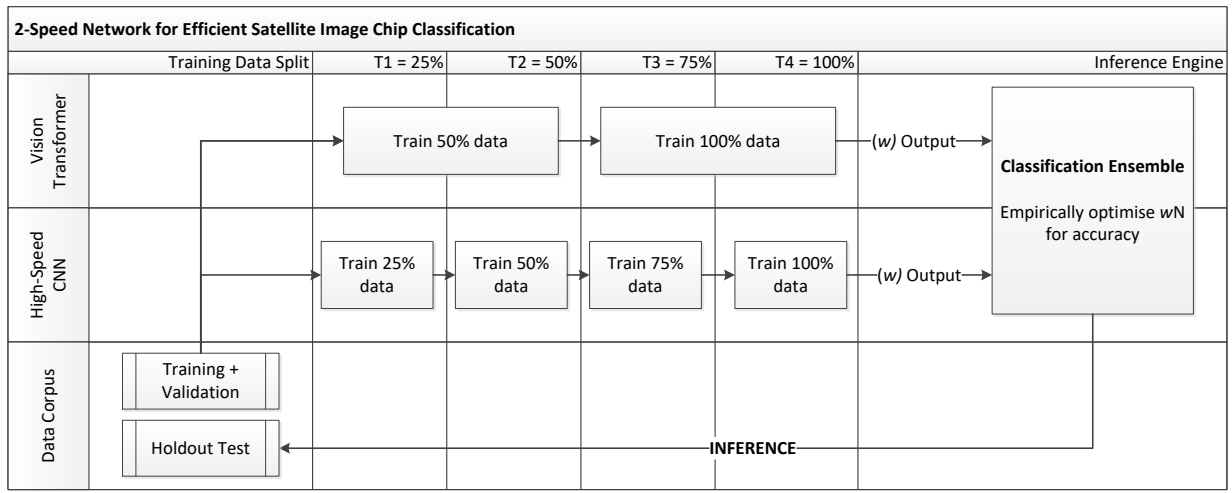

**Figure 1** – 2-Speed network ensemble process flow for efficient satellite image chip classification using staggered training over four simulated time intervals.

## 2.1 Dataset details

The So2Sat dataset consists of image chips acquired from the Sentinel-1 and Sentinel-2 missions with 17 local climate zone (LCZ) labels assigned by a team using a rigorous workflow including peer verification and quantitative evaluation resulting in a claimed label confidence of 85% [26]. This dataset consists of 400,673 such image chips from 42 cities at 10m resolution. Each image chip measures only 32x32 pixels resulting in a small and pixelated appearance of sample images. However, the So2Sat dataset is an order of magnitude larger than any other publicly available satellite image dataset. Since our investigation is primarily concerned with the adaptability and scalability of deep learning models in the context of incremental data, So2Sat is the only suitable satellite image dataset for this study.

The So2Sat authors established a baseline classification overall accuracy for several machine learning algorithms including random forest (RF), support vector machines, and attention augmented variation of ResNeXt [33]. The best overall accuracy in this source paper was 0.61 achieved using the ResNeXt based classifier. The overall accuracy metrics of 51% and 54% were achieved by RF and SVM respectively. The small size of the So2Sat image chips results in relatively low classification accuracy in studies using this dataset, compared to studies using higher resolution satellite images. For example, state-of-the-art classification for the high-resolution AID [28] and NWPU-45 [12] datasets has recently been reported with overall accuracy of 97.01% and 94.46% respectively [34] using a dual branch network comprising a pair of ResNet [35] CNNs.

Examples for each of the 17 labels in this dataset are shown in Figure 2.

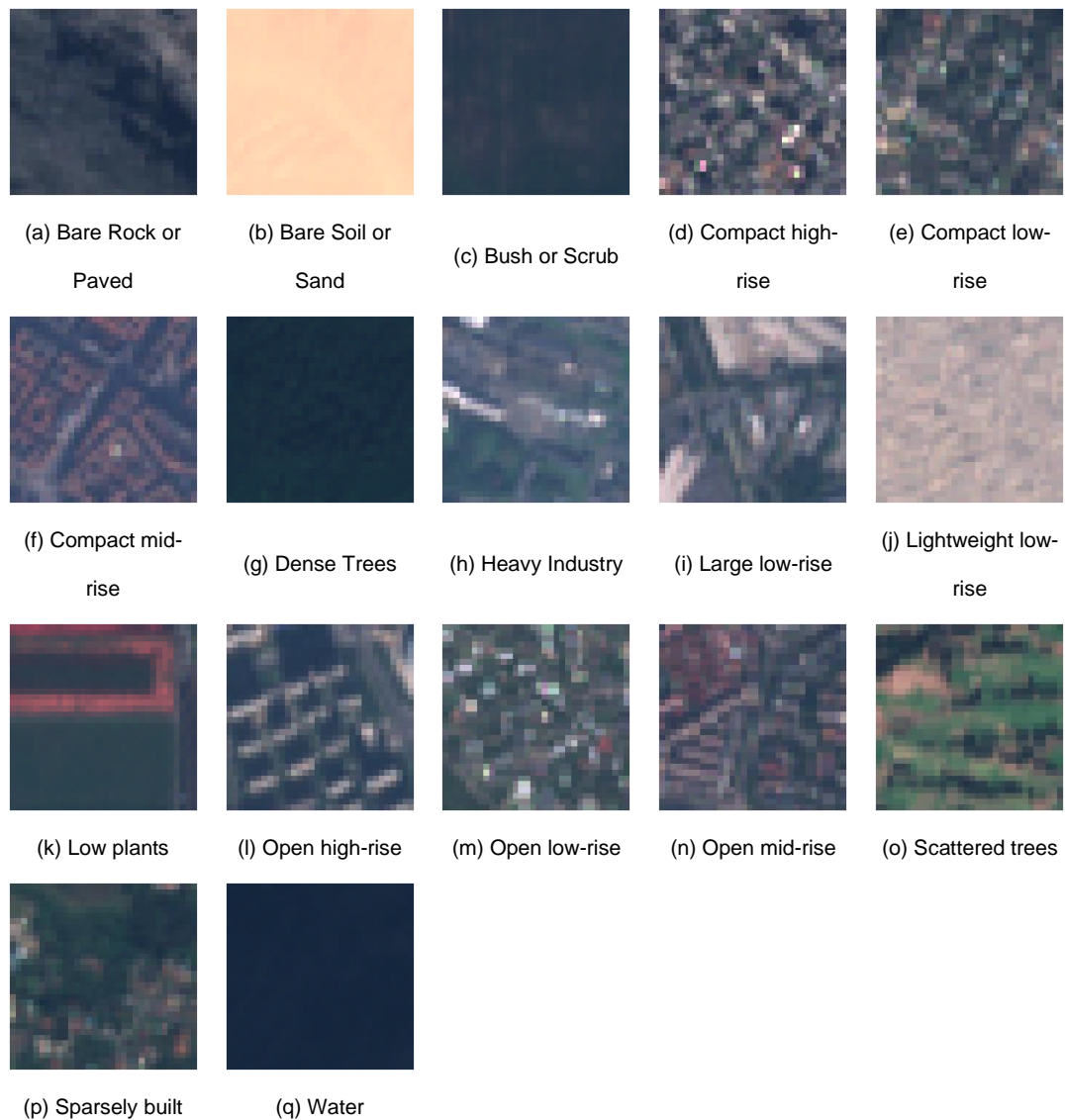

**Figure 2 -** So2Sat samples showing 17 land-use/land-cover types (using standardized LCZ labels) with a mix of local and global features.

A recent benchmarking study on the multi-band version of this dataset by [36] achieved the best overall accuracy of 69% using a complex multi-level fusion CNN with 16 filters width of the first block and network depth of 17 layers. This architecture had the novelty of globally averaging each layer's maximum average pooling output as input to the 17-layer dense output head. The overall accuracy benchmark for the RGB version of the So2Sat dataset that we are using in this study is 61% as reported by [26]. We have used the RGB version of the So2Sat dataset in this study primarily to allow for the use of out-of-the-box three-channel image classification and thereby allow for direct comparison with the benchmark study.

**2.2 Network Architectures**

We have chosen two distinct network architectures as complimentary pairs for this study. Firstly, a 16 patch Vision Transformer [23] (ViT) network architecture has been selected as a highly accurate and scalable image classification network. Secondly, a High-Speed Convolutional Neural Network (HS-CNN) [11] architecture has been handcrafted as a low parameter image classifier for high-speed training ideal for the incorporation of new data into a computer vision model.

ViT based architectures have recently come to dominate ImageNet [37] computer vision classification with state-of-the-art accuracy emerging from several studies [23, 38, 39]. ViT architectures interpret images as sequences of patches with each patch processed by multi-head/multi-layer perceptron (MLP) attention layers to generate a feature space that is then classified by a densely connected softmax output layer.

CNNs tend to provide excellent performance on small to medium-sized datasets due to the relative ease with which CNNs derive inductive biases by automated feature extraction. For larger datasets, the scalability of the ViT architecture "trumps" the inductive bias advantage of CNNs resulting in better classification performance at a large scale [23]. For this reason, ViT architectures have recently been proven highly effective in remote sensing applications using satellite imagery achieving state-of-the-art [40] results across four datasets, UC-Merced [27], AID [28], Optimal31 [29], and NWPU-45 [12]. An obvious advantage of the ViT architecture in remote sensing applications is the faster training over large datasets compared to very deep CNNs of around 80% compared to the EfficientNet architecture. For this study, we have selected a 16 patch ViT architecture uses 12 encoder layers, a hidden size of 768, MLP size of 3072, with 12 self-attention heads resulting in a model with 85.7 million trainable parameters. This ViT architecture variant has been selected as the smallest and, therefore, least training resource-intensive option given the small size of the So2Sat image chips (32x32). The ViT was pre-trained on ImageNet classes and shared with the community [41].

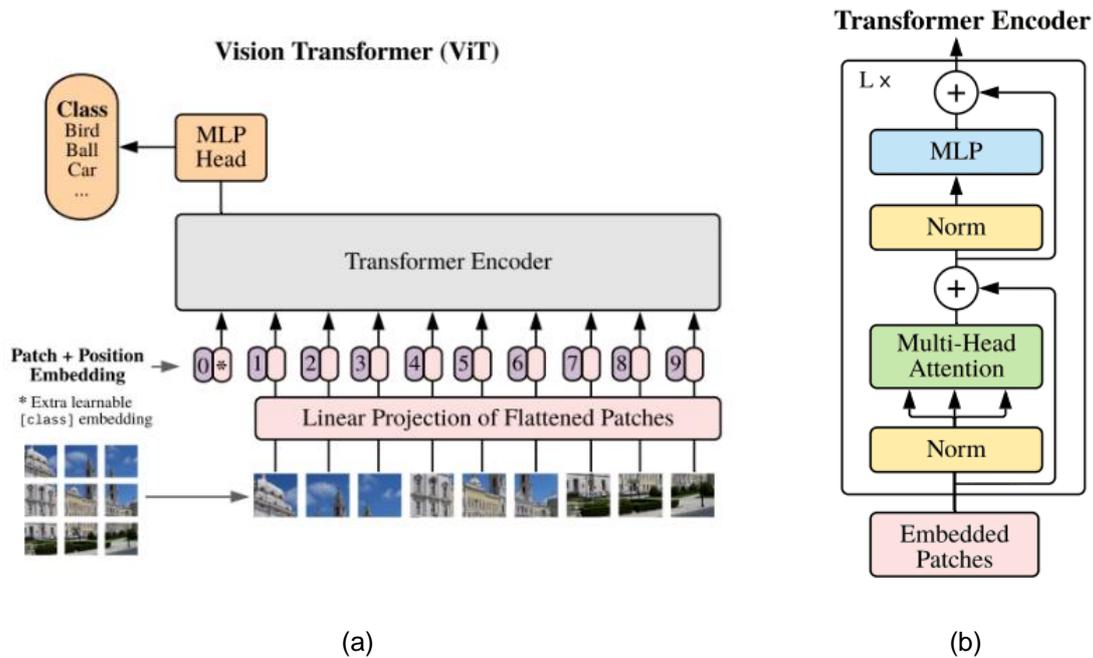

**Figure 3** - ViT architecture schematics from [23] showing (a) high level transformer architecture and (b) detail of the transformer encoder. In this study, we used the base ViT layout with 12 MLP heads and 12 transformer encoder layers (L = 12).

The ViT architecture is presented in Figure 2(a). An image is first split into patches (16 for our study) before being flattened and embedded as lower-dimensional representations along with the patch position in the sequence. This combination of embedded image and position is fed in sequence to a transformer encoder consisting of multiple transformer encoder blocks. Each encoder block comprises a multi-head self-attention (MHA) layer followed by normalization and a MLP head with short residual skip connections used to allow representations from earlier layers to interact with later layers. Essentially, the ViT allows for computer vision to learn image classification by both pixel values and the positional context of the image patches.

Although the ViT architecture is both efficient and scalable, it requires a large number of samples before overtaking traditional CNNs in terms of classification metrics [23]. In the case of new data added to an already large data corpus, complete retraining of the ViT will be still expensive and time-consuming, despite the efficiency advantages of the ViT architecture over CNNs. For this reason, we have handcrafted a low parameter high-speed CNN (HS-CNN) classifier derived from the VGG architecture [14] but with only three layers, each comprising two convolution layers. This network has been designed with the objective to minimize training time while maintaining good accuracy for classifying new data.

The number of trainable parameters for the HS-CNN is 2.8M. For comparison, other commonly used CNN architectures such as VGG16 [14], ResNet18 [35], and ResNeXt [33] have 138.4M, 11.5M, and 25M trainable parameters, respectively. To minimize overfitting to this very sparse network architecture, each max pooling layer is followed by a dropout layer, and fully connected layers were regularized using an L2 regularization penalty [42]. The HS-CNN architecture used in this paper is shown in Figure 3. The HS-CNN is initialized with random weights and biases and trained from scratch using the So2Sat image chip dataset.

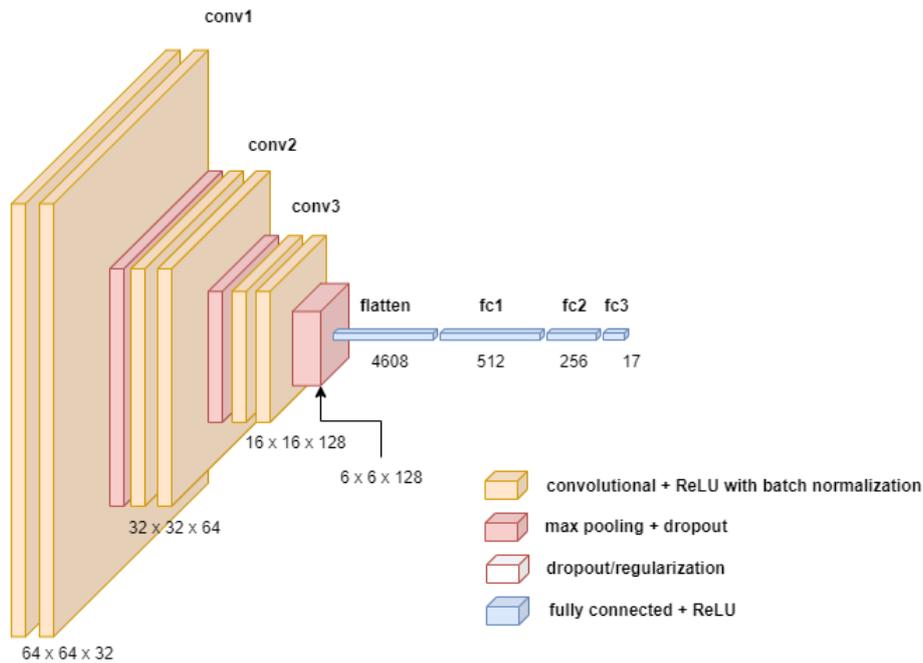

**Figure 4** – HS-CNN architecture overview showing a VGG-like structure with only 3 layers and additional regularization designed to minimize training time whilst avoiding overfitting.

**2.3 Ensemble Architecture**

Since the ViT breaks an image into patches (16 for this study) and then encodes each patch with positional embedding as an input to the transformer encoder, the ViT learns global features of an image simultaneously with pixel values [43]. In contrast, since a CNN is trained by learning the co-relationships of overlapping small arrays of pixels, the CNN learns pixel-based local features first, with long-range global features becoming emergent as training proceeds.

We hypothesize that the contrasting learning strategies modes of ViT and HS-CNN make these models good candidate components for an ensemble model (confirmed in results section 3.3), whereby outputs from each component model are combined via a weighted averaging algorithm to arrive at a final prediction according to equation 1.

$$\bar{p} = \frac{\sum_{i=1}^{n} pi \times wi}{\sum_{i=1}^{n} wi} \qquad (1)$$

Here, $p$ is the predicted score for samples from the $i$ classifier and $w$ is the weight assigned to predictions from that classifier. In this study, each classifier has been assigned a weight ranging from 0.1 to 0.9 with steps of 0.1 with each classifier's weights adding to unity on each test. An industrial implementation of the proposed staggered learning scheme would include the classifier weights as a learnable parameter to automatically optimize the predictive value of the ensemble.

**2.4 Staggered Training Schedule**

Since this study is concerned with additional data at four points in time, four classification models are used in testing as described in table 2. CNN-25 at initial time T1 is a HS-CNN trained and validated on 25% of the data and representing a point in time (T2) where there has been sufficient time to train the HS-CNN but not the ViT. ENS-50 represents the point in time (T3) where the ViT has completed training on 50% of the data along with the CNN also having been trained on 50% of the data. ENS-75 represents a point in time partway through the next ViT training cycle where the ViT model is still only available as trained on 50% of the data, but the high-speed CNN has been trained on 75% of the data. Finally, ENS-100 represents a point in time (T4) where both classifiers are fully trained on 100% of the data. A summary of the staggered training schedule and model naming convention used in this study appears in Table 2.

**Table 2.** A staggered training schedule was used to mimic the availability of new data at four points in time denoted as T1 to T4. The high-speed CNN is trained every increment. The ViT is trained every 2 increments. Ensembles are created at each increment using the most recently trained component model.

| Model Name | Time Increment | Training Data | CNN Training | ViT Training |
|---|---|---|---|---|
| CNN-25 | T1 | 25% | 25% | None |
| ENS-50 | T2 | 50% | 50% | 50% |
| ENS-75 | T3 | 75% | 75% | 50% |
| ENS-100 | T4 | 100% | 100% | 100% |

**2.5 Experiment Setup with Incremental Data Simulation**

The So2Sat corpus is available as a TensorFlow dataset [44] providing both multi-channel and JPEG encoded reg, green, and blue (RGB) images. For this study, we have selected the RGB subset to allow for a fair comparison to the deep learning classifier results from the So2Sat source paper [26] and also to generalize the potential merits of the approach to other three-channel, visible spectrum computer vision tasks. The So2Sat dataset includes a standard split for model training and testing purposes. This split provides a total of 352,366 images for training/validation and 24,119 for holdout testing. Each model was trained and validated on increasing 25% increments of the training data but tested against the entire holdout testing corpus in order to provide a fair comparison of predictive capability at each simulated time increment. Training data was augmented with random left/right/up/down flipping along with random brightness, contrast, and saturation operations. Testing data was not augmented in any way. All images were shuffled before being used to train/test classifiers to eliminate sampling biases that may have been caused by data collection order, for example local geographical confounders such as regional standard for roofing materials, building and industrial layouts.

**2.6 Compute Configuration**

All experiments were executed on the University of Technology Sydney Interactive High Performance Compute environment hardware and software as described in Table 3.

**Table 3.** Summary of hardware and software configuration for this study.

| Hardware | Software |
|---|---|
| CPU: Intel Xeon E-2288G @ 3.70 GHz | OS: Red Hat Enterprise Linux v7.9 |
| Memory: 64GiB | Python: 3.8.12 |
| GPU: Nvidia Quadro RTX 6000 with CUDA v 11.2 | TensorFlow: 2.7.0 |

**3. Results**

*3.1. Model Training and Validation*

Training curves for the ViT and HS-CNN classifiers, when trained against the complete So2Sat training data set, are presented in Figures 5 and 6. The ViT classifier training chart shows good convergence without overfitting with excellent validation accuracy of 0.92. The HS-CNN training curve also shows good convergence, especially for a scratch-trained network, but with an overall lower validation accuracy of 0.82. Training behavior was essentially identical regardless of the data split used, with the only noticeable difference being a slower convergence for the ViT classifier when trained with the 25% split.

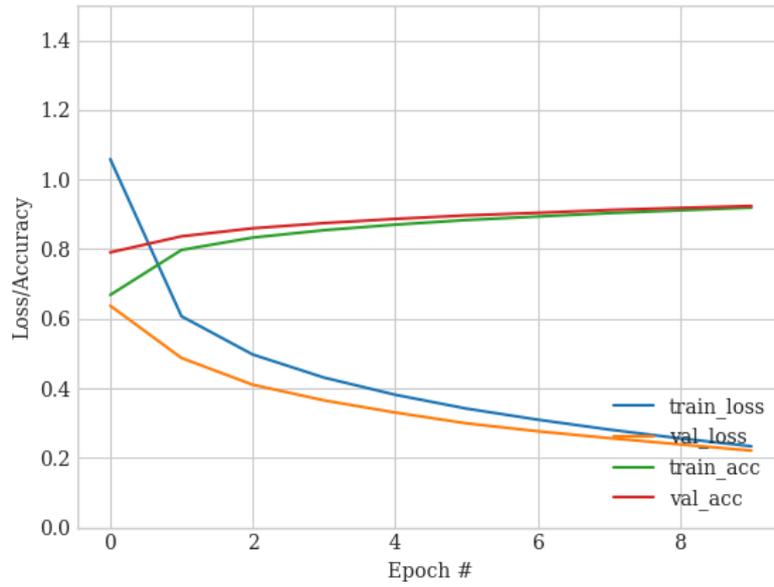

**Figure 5** – ViT training curve. Trained for 10 Epochs at a learning rate of 1e-6. The model converged well resulting in very high validation accuracy of 0.92 when trained on the So2Sat full training dataset.

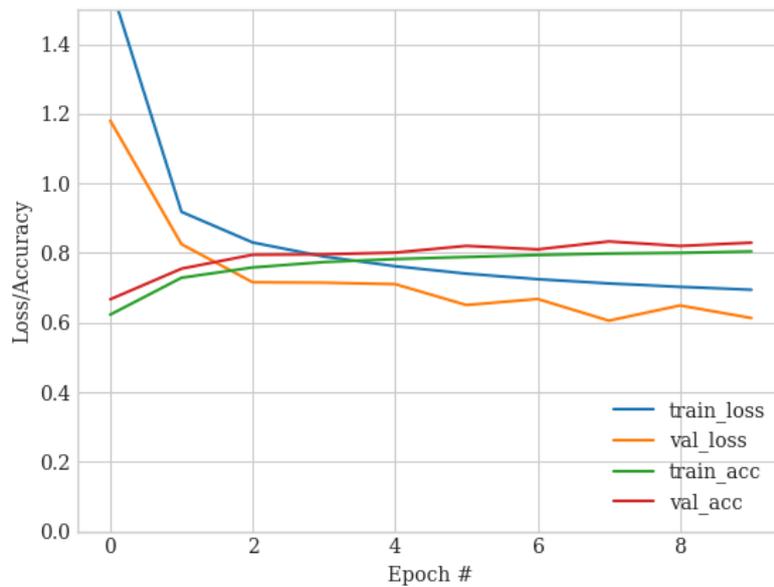

**Figure 6** – High-speed CNN training curve. Trained for 10 Epochs at a learning rate of 1e-4. The model converged quickly reaching a validation accuracy of 0.82 when trained on the So2Sat full training dataset.

*3.2. ViT and HS-CNN Training Metrics and Holdout Testing*

We were interested in whether the very high validation metrics achieved during training would generalize to the holdout test dataset. Our results showed that the clean model training did not translate entirely to holdout testing metrics. The HS-CNN achieved a best holdout test overall accuracy of 0.61 when using 25% of the training data and 0.60 when using the full training data set. This is lower than the benchmark of 0.61 set by [26] using complex attention augmented ResNeXt architecture but still a reasonable result given that the HS-CNN has an order of magnitude less training parameters than the ResNeXt (2.8M vs ~25M) [33] architecture used in that study. The HS-CNN meets its design objective of good accuracy at high training speed, with the full training dataset of 317,129 image chips processed in 28 minutes. As expected, the ViT showed improved performance over the HS-CNN with peak overall accuracy of 0.63 using 50% of the training data, dropping to 0.62 when 100% of the training data was used. The ViT performance represents a marginal improvement on the benchmark overall accuracy of 0.61. The ViT took over 4 hours to train with the full training set, which is over 8 times the training time of the HS-CNN.

**Table 4.** Training metrics summary for each data increment. Holdout test results were obtained against the full So2Sat test set of 24,119 images.

| Classifier | Training Split (%) | Training Image Count | Validation Image Count | Training Time (h:mm:ss) | Validation OA | Holdout Test OA |
|---|---|---|---|---|---|---|
| HS-CNN | 25 | 81,044 | 7,048 | 0:07:23 | 0.78 | 0.61 |
|  | 50 | 158,565 | 17,618 | 0:14:36 | 0.81 | 0.60 |
|  | 75 | 239,609 | 24,665 | 0:21:35 | 0.81 | 0.58 |
|  | 100 | 317,129 | 35,237 | 0:28:34 | 0.82 | 0.60 |
| ViT | 25 | 81,044 | 7,048 | 1:03:45 | 0.82 | 0.63 |
|  | 50 | 158,565 | 17,618 | 2:04:30 | 0.89 | 0.63 |
|  | 75 | 239,609 | 24,665 | 3:07:29 | 0.91 | 0.62 |
|  | 100 | 317,129 | 35,237 | 4:09:39 | 0.92 | 0.62 |

*3.3. Ensemble Model Holdout Testing Results*

Three ensemble models were created using variously trained HS-CNN and ViT models as follows:

- ENS-50 consisting of the HS-CNN and the ViT each trained on 50% of the training data,
- ENS-75 consisting of the HS-CNN trained on 75% of the training data and the ViT trained on 50% of the training data, and
- ENS-100 consisting of the HS-CNN and the ViT each trained on 100% of the training data.

*3.3. Ensemble Model Holdout Testing Results*

Results of holdout testing for the ensemble models at each time increment are presented in table 5. At time increment T1 using 25% of the training dataset partition, the only trained model is the HS-CNN. Therefore, results are identical to those obtained using HS-CNN at a 25% training split. For time increment T2, the ViT and HS-CNN are both trained using 50% training data. At time T3 the ViT and HS-CNN are trained on 50% and 75% of training data, respectively. At T4 both the ViT and the HS-CNN are trained on 100% of training data. This allows the HS-CNN and ViT to be ensembled with predictions used as inputs to the weighted averaging function described in equation 1. A scripted experiment varied the HS-CNN:ViT weighting by 10% from 10:90 to 90:10. Best, and identical, results were achieved using ENS-75 with weighting ratios of 40:60, 50:50, and 60:40 as shown in Figure 7.

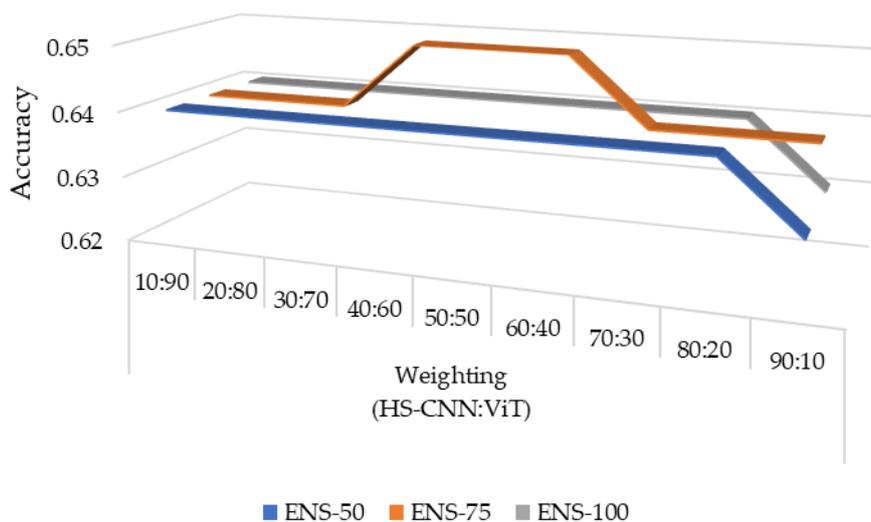

**Figure 7** – Ensemble weighting results. Effect of different weight ratios for ensemble component models. Best accuracy is achieved using a balanced HS-CNN:ViT weighting ranging from 40:60 to 60:50

The result of holdout testing including the ensemble models is provided in table 5. In general, the overall accuracy results at times T2, T3, and T4 given by the ensemble models are better than those of either component model at the same data partition. ENS-50 comprising HS-CNN trained on 50% data and ViT also trained on 50% data achieved holdout test an overall accuracy of 64%. The highest overall accuracy was achieved by ENS-75 comprising HS-CNN trained with 75% data and ViT trained on 50% data. This ensemble achieved an overall accuracy of 65%, which is a substantial improvement over the baseline overall accuracy of 61% [26]. ENS-100, consisting of an ensemble of fully trained HS-CNN and ViT achieved an overall accuracy of 64%. Precision and recall metrics were well balanced for all tests indicating that the accuracy has not been achieved through simple over-classification of majority classes.

**Table 5.** Results of inference for each staggered training time interval. Holdout test results were obtained against the full So2Sat test set of 24,119 images. Ensembles were weighted 50:50 for each classifier.

| Model Name | Time Increment | Total Training Time | Holdout Test Precision | Holdout Test Recall | Holdout Test F1 | Holdout Test OA |
|---|---|---|---|---|---|---|
| CNN-25 | T1 | 0:07:23 | 0.59 | 0.61 | 0.58 | 0.61 |
| ENS-50 | T2 | 2:19:06 | 0.63 | 0.64 | 0.62 | 0.64 |
| ENS-75 | T3 | 2:29:05 | 0.64 | 0.65 | 0.63 | 0.65 |
| ENS-100 | T4 | 4:38:03 | 0.64 | 0.64 | 0.63 | 0.64 |

To illustrate the effectiveness of the ensemble approach in relation to both accuracy and efficiency, Figure 7 presents comparative plots of all tested models. Figure 7(a) shows that the ensemble models improve accuracy over component models for all models at all data partitions. In addition, the ensemble ENS-75 provided the highest accuracy of all tests with training time being approximately half that of the fully trained ViT model as shown in Figure 7(b).

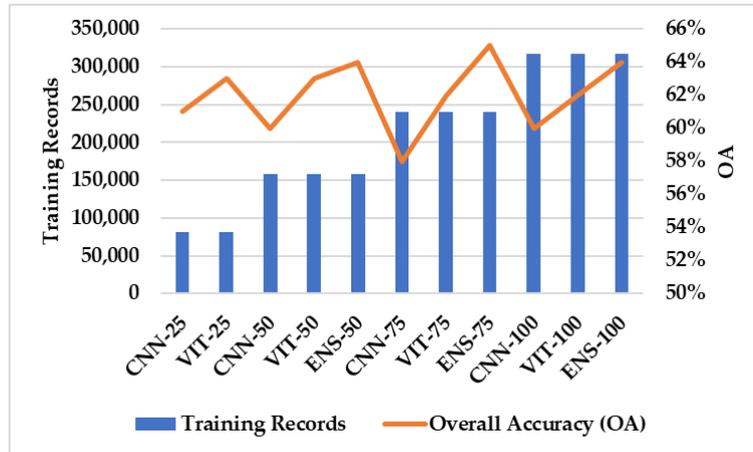

(a)

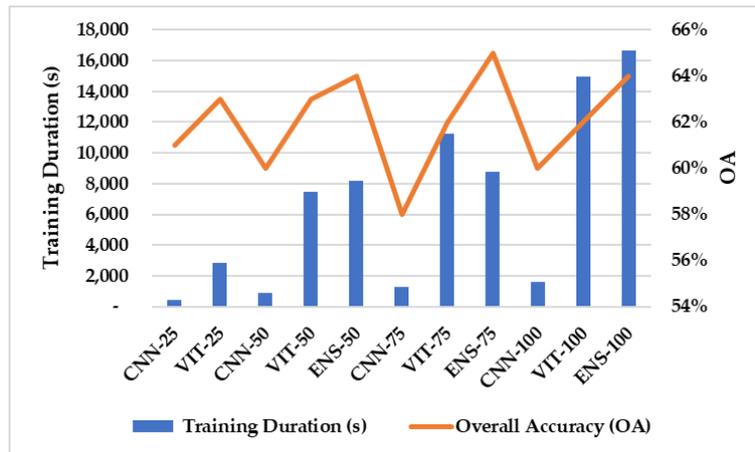

(b)

**Figure 7** – Comparison of holdout test accuracy results for all models including HS-CNN, ViT and ensembles. (a) is Classification Accuracy by Training Data Size. Ensemble models composed of ViT and Lightweight CNN show higher accuracy with less training data than ViT or CNN trained on larger datasets. (b) is Classification Accuracy by Training Duration. Ensemble model consisting of ViT trained on 50% data and Lightweight CNN trained on 75% of data provides the best accuracy of 65% with training time approx. 40% lower than a fully trained ViT.

*3.3. Ensemble Model Classification Analysis using Confusion Matrices*

Confusion matrices associated with each classifier and the ensemble at training interval T3 were generated to analyze the relative strengths of each approach. The confusion matrices for HS-CNN, ViT, and ENS-100 are shown in Figures 8(a), 8(b), and 8(c), respectively.

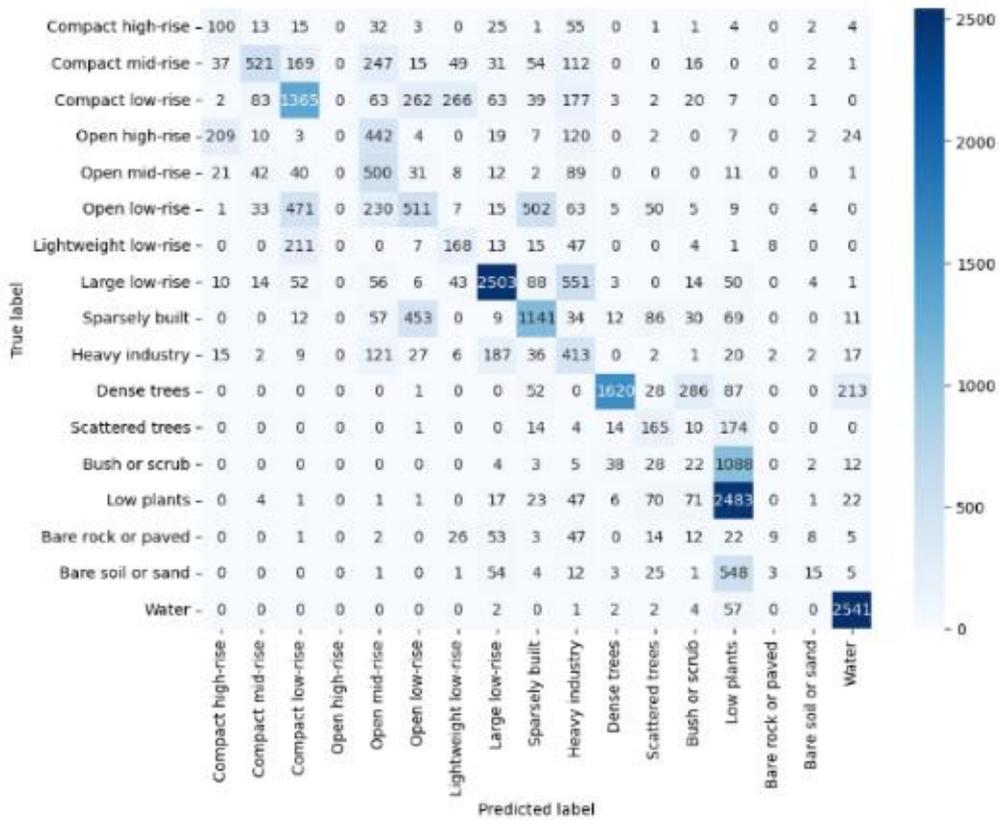

(a)

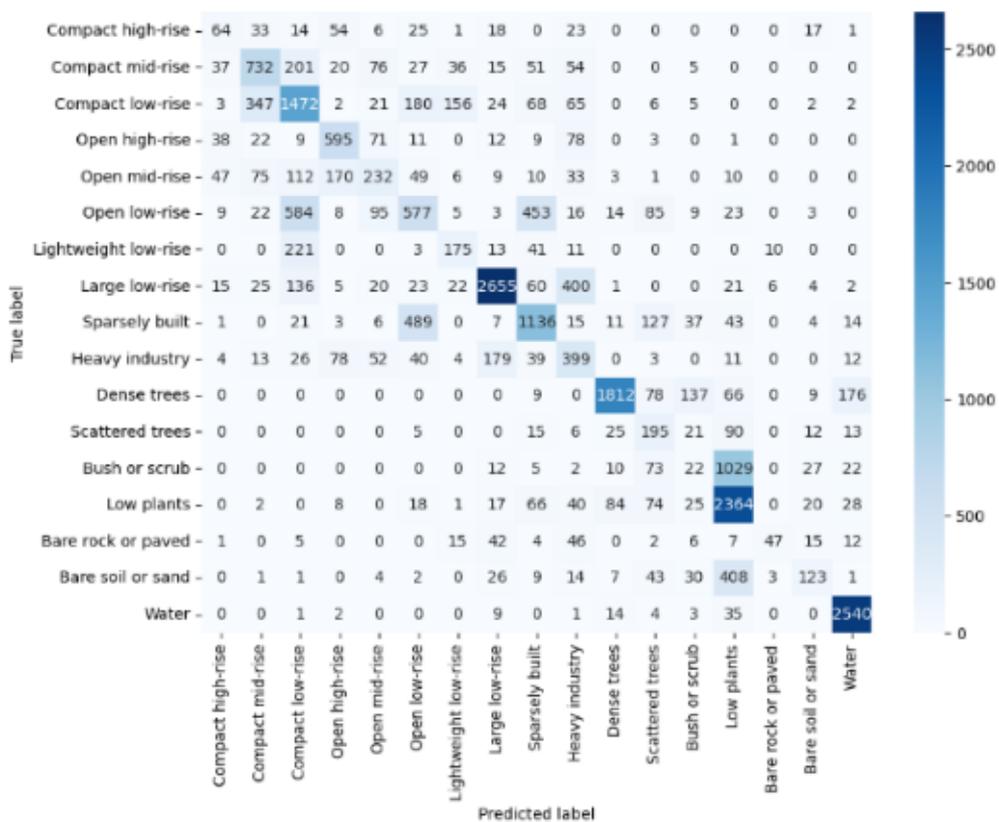

(b)

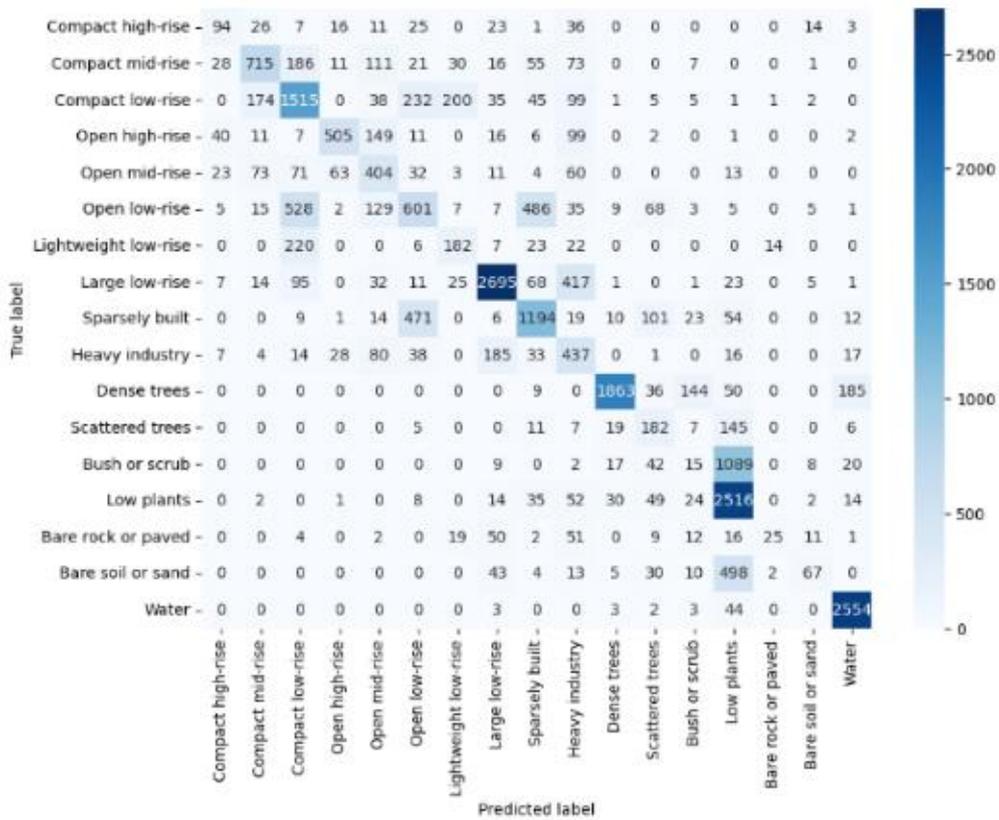

(c)

**Figure 8.** Confusion matrices relating to the best ensemble model ENS-75. (**a**) Lightweight CNN trained on 75% of data. (**b**) 16 patch ViT using 50% of training data. (**c**) Ensemble model ENS-75 taken at T3.

As the ViT achieved higher overall accuracy than the HS-CNN we first consider the class labels contributing most to this accuracy delta. The top three such classes are "Open high-rise", "Bare rock or paved", and "Bare soil or sand". The HS-CNN failed to classify any "Open high-rise" correctly, and instead classified the majority (n=442) of true "Open high-rise" as "Open mid-rise". Recalling Figures 2(l) and 2(n) as examples of these two classes, the "Open high-rise" examples show regular building alignments that are not present in the open mid-rise. The pattern of these regular alignments is apparent as long-range diagonal features, explaining the ViT superior performance in classifying these classes. Similarly, the ViT outperformed the HS-CNN in separating the classes with sparse local features such as the "Bare" classes in Figures 2(a) and 2(b), where the visible features are long range features across the image chip such as topographical features in the case of bare rock or paved, or sand dune formations in bare soil or sand. The two classifier types provide similar performance for chip classes that lack long-range features such as water, dense trees, and bush or scrub, as evident from the confusion matrices in Figure 8.

To illustrate this class activation maps have been generated for these divergent classes are shown in Figure 9 (a-g). Figure 9(b) illustrates the ViT attention on the long-range feature of building alignments for the Open high-rise class whereas the HS-CNN in Figure 9(c) attends to a less focused regions of pixels that are a mix of buildings and open space, resulting in the HS-CNN proving to be unable to distinguish between Open high-rise and Compact high-rise, Open mid-rise, and Heavy Industry. In a similar manner, Figure 9(e) illustrates the ViT attending to the bare rock feature in the upper right corner of the image chip, which is an area of low attention to the HS-CNN 9(f). Finally, the ViT appears to have identified sand dune areas in Figure 9(h) with the HS-CNN failing to attend to any feature at all – to the HS-CNN the "Bare soil or sand" image chip is featureless since it is poor at identifying the long-range sand dune edges when compared to the ViT.

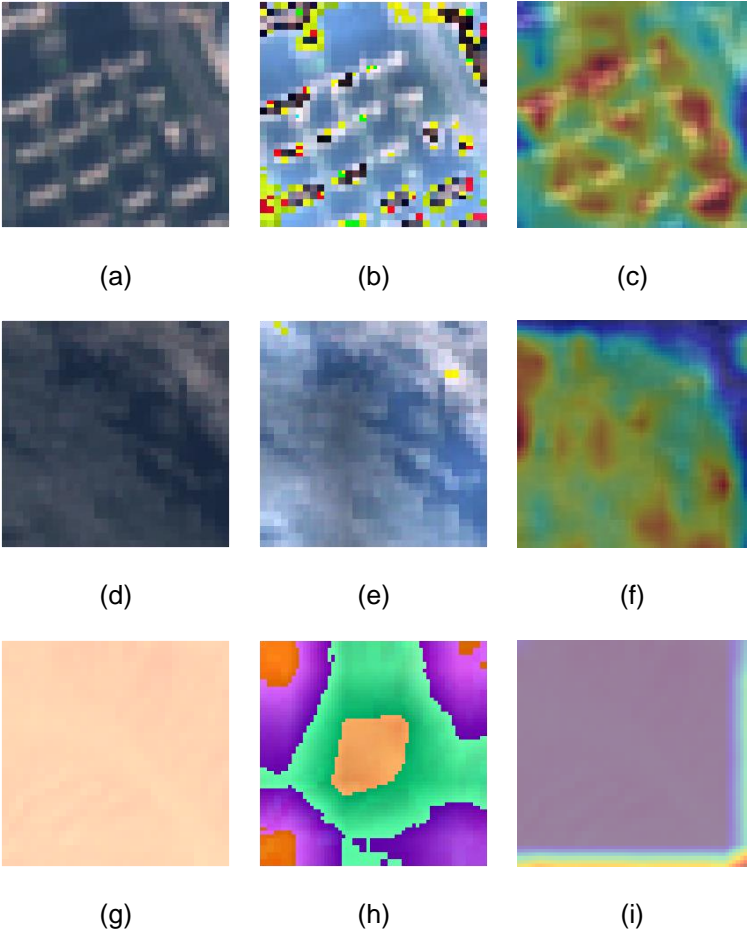

(a) (b) (c)

(d) (e) (f)

(g) (h) (i)

**Figure 9.** ViT activation maps for class labels more accurately separated by the ViT over the HS-CNN. (a) Open high-rise with (b) Open high-rise ViT activation map accurately tracking the building alignment, and (c) Open high-rise HS-CNN activation map also tracking the building alignment but at much lower resolution. (d) Bare rock or paved with (e) ViT activations tracking long-range topographical features as a feature in the top right corner, and (f)

HS-CNN also tracking some topographical features but with poor resolution resulting in widespread activation over the chip and no attention to the bare rock in the top right corner, (g) Bare soil or sand with (h) ViT activations again tracking the long-range sand dune edges, and (i) HS-CNN with no relevant activations for this chip.

**4. Discussion**

Incorporating new data into artificial intelligence vision systems will remain a challenging problem, since complete re-training of these systems is resource-intensive, and other techniques such as teacher-student models and fine-tuning with new data can also be problematic. Increases in computing power over time, particularly GPU processing, are quickly consumed by the desire to train deep learning systems on more significant numbers of higher-definition images, thereby instantly consuming compute improvements. Using a 2-speed ensemble network comprising a HS-CNN combined with a slower but more accurate Vision Transformer network provides a practical means of dealing with incremental changes, including new labels, to a very large dataset by regular retraining. The ViT provides accuracy and scalability but with high compute resource costs incurred in training. The HS-CNN provides adaptability to new data with meagre compute resource cost and very fast training times. The complimentary resource profiles of these architectures allow for an optimized training schedule whereby the HS-CNN is trained at a higher frequency than the ViT. In our experiments, the combination of staggered training with an ensemble model led to state-of-the-art results for classification of the So2Sat dataset with accuracy of 65% achieved in holdout testing, using a fully trained HS-CNN and a ViT trained on only 75% of the available data corpus. This result improves on the previous overall accuracy baseline of 61% and is, to the best of our knowledge, current state-of-the-art for the RGB version of the So2Sat dataset.

Although the initial objective of this study was to improve efficiency of image chip classification for very large datasets in the presence of new data, the underlying contributors to our results are worthy of investigation. Labels that were better separated by the ViT over the HS-CNN were identified, with network attention maps indicating that the ViT is superior to the HS-CNN in detection of long-range features, even in the small So2Sat image samples where such features are limited to 10m. This is because the ViT includes chip patch position as training input, whereas the HS-CNN training input is limited to localized pixel arrays. Therefore, the ViT can better train on features that span the image chip, such as building alignments and topographical features.

## 5. Conclusion

Efficient, automated processing of incremental big data will remain an open problem due to exponential growth in data outpacing linear growth in compute resource efficiency and availability. We have shown that the use of complimentary computer vision algorithms along with a staggered training schedule results in efficient training, and overall accuracy is greater than that achieved by each individual algorithm against the same dataset. This 2-speed network reduces the number of full data corpus training cycles in favor of the creation of an agile ensemble model with lower overall training time needed to reach the best accuracy. This approach also allows model production release to be undertaken using an agile software methodology/pipeline, whereby the expensive but more accurate ViT model is considered a major release, with revised HS-CNN models considered to be a point release.

In the future, we intend to progress this idea to an industrial trial whereby empirical performance data can be used to tune the ViT and HS-CNN architectures, hyperparameters, and prediction weightings resulting in a domain-specific ensemble that is efficient to train and adaptable to new data. This will provide a valuable tool for strategic planning agencies to formulate actions in response to changes in the landscape. Finally, we are currently investigating enhancing the 2-speed network ensemble by the inclusion of a few-shot learning engine based on an edge-labelling graph neural network as suggested by [45] as a means of adding a real-time classification capability for unseen images.

**Funding:** This research was supported by Defence Australia funding under the AI for Decision Making Initiative project titled "Effective updating of deep learning models with limited new data" (UTS Ref: 210018980).

**Acknowledgments:** The first author would like to acknowledge their employer, IBM Australia Limited for providing writing time to produce this manuscript.

**Declaration of Interest:** The authors declare no conflict of interest. The funders had no role in the design of the study; in the collection, analyses, or interpretation of data; in the writing of the manuscript, or in the decision to publish the results.